# A new approach for digit recognition based on hand gesture analysis

Ahmed BEN JMAA[#1], Walid MAHDI[#2], Yousra BEN JEMAA[*4] and Abdelmajid BEN HMADOU[#3]

\# Multimedia InfoRmation systems and Advanced Computing Laboratory

Higher Institute of Computer Science and Multimedia

Sfax, Tunisia

[1]ahmed.benjmaa@gmail.com

[2]walid.mahdi@isimsf.rnu.tn

[3]abdelmajid.benhamadou@isimsf.rnu.tn

\* Signal and System Research Unit

National Engineering School

Tunis, Tunisia

[4]yousra.benjemaa@planet.tn

*Abstract—* **We present in this paper a new approach for hand gesture analysis that allows digit recognition. The analysis is based on extracting a set of features from a hand image and then combining them by using an induction graph. The most important features we extract from each image are the fingers locations, their heights and the distance between each pair of fingers. Our approach consists of three steps: (i) Hand detection and localization, (ii) fingers extraction and (iii) features identification and combination to digit recognition. Each input image is assumed to contain only one person, thus we apply a fuzzy classifier to identify the skin pixels. In the finger extraction step, we attempt to remove all the hand components except the fingers, this process is based on the hand anatomy properties. The final step consists on representing histogram of the detected fingers in order to extract features that will be used for digit recognition. The approach is invariant to scale, rotation and translation of the hand. Some experiments have been undertaken to show the effectiveness of the proposed approach.**

## I. INTRODUCTION

The gesture is a natural way of communication for many deaf with sign language. Many machines are designed to be operated by gestures through mechanical and electronic interfaces, such as a driving car, a robot remote control…

The gesture recognition is very difficult and complex task since the full recognition system should be able to identify the hand in different scales, positions, orientations, contrasts, funds, Luminosity, and others.

However, many methods, has been developed in hand detection, classified into many categories: knowledge-based methods, feature based methods, template-based methods, and appearance based methods [16][17][18][19][20].

When used separately, these methods cannot solve all problems of hand detection. Hence, it is better to operate with several successive or parallel methods.

The gesture communication with machines are enriched by instrumented gloves [11], which are used to control a virtual actors, to describe and manipulate objects on computer screens [1][2] or even to recognize the sign language [3]. Unfortunately, these gloves are expensive and fragile, and their cables are a hindrance. Thus many researches are rather interested in gesture acquisition. The hand gesture acquisition with cameras for "Human-Machine" interaction applications or sign language recognition need to localize the hand in the picture, then to evaluate the 2D settings such as fingers tips positions or articulation angles. Gesture analysis can make use of colored gloves or markers on the hand [4]. The 2D analysis methods of gestures are limited to recognize a limited number of predefined layouts. Berard et al. [5] have developed a drawing application that provides a real-time position of a finger, by correlation on the fingertip. Cootes et al. [6] have proposed deformable models by using a model of distribution points, which represents any form of skeleton by a set of feature points and variation patterns that describe the movements of these points. This model is made from a set of training sequence by a singular values decomposition of the differences between the forms of a set of training sequence and the averaged form. The recognition is based on the model of distribution points. This has also been proposed by Martin and Crowley [7].

As for us, we consider that the fingers localization provides a powerful issue of evidence to predict the hand sign. Thus, fingers analysis could be used efficiently to hand gesture recognition.

In this paper, we present a new hand detection technique based on the color classification using fuzzy logic system, image segmentation using edge detection and we try testing two methods (Ellipse method and Comparative based method) in order to localize the hands from others objects detected in the image. The results obtained from this new technique of hand detection and hand localization are used in our hand gesture recognition systems.

Then, we present a method that put the fingers into a uniform localization. Based on its properties, the fingers pixels distribution are basically the main input parameters to the gesture recognition process. We are based on digit recognition which represents a meaningful need in the sign language framework, to show the effectiveness of our hypothesis.

This paper is organized as follows, in section two we describe the hand detection system and adjustment of its orientation. In section three we present all techniques that we use to remove all hand components except the fingers. In section four the digit system recognition will be described. Finally, we present some experiments and results to show the effectiveness of the proposed approach.

## II. HAND DETECTION AND LOCALIZATION

Hand detection is a very important stage to start the step of sign language recognition. In fact, to identify a geste, it is necessary to localize hands and their characteristics in the image. Our system supposes that only one person is presented in the field of the camera, it is described in (figure 1).

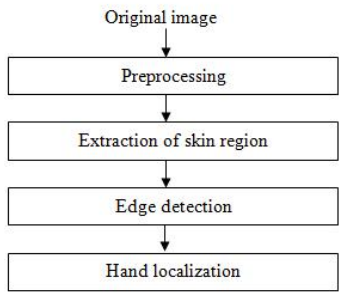

Figure 1.   Steps of hand detection

### A.   Skin color detection

To improve the quality of the image, we attenuate the noise by applying a low-pass filter. Many techniques have been proposed to determine human skin color and results are usually submitted to conditions of lighting [21][22][23].

### 1)   Skin color space:
It is known that human skin color contains a majority of red color due to the blood, but this remains largely insufficient to develop a robust algorithm using the *RGB* space. In fact, the values of *R*, *G* and *B* component are depending on the lighting conditions. We propose for our system to use the *YCbCr* color space in order to dissociate the luminance from the color information. Furthermore, the *Cb* and *Cr* are the chrominance components used in MPEG and JPEG [9]. In (figure 2), we illustrate how the *YCbCr* space can separate the skin color from the background.

In order to classify each pixel of the image in skin pixel and non skin pixel, the most suitable arrangements that we found for all images are *Cb* in [77, 127] and *Cr* in [139, 210]. These terminals are approved experimentally and are independent of race human skin color [10].

### 2)   Fuzzy classification:
The proposed arrangements are not sufficient to find a good classification because of the diversity of human skin color and for many other reasons such as noise and shad. To overcome the limitation of classic classification, we propose to apply a fuzzy approach for pixel classification.

This is considered as a good solution since that, fuzzy set theory can represent and manipulate uncertainly and ambiguity [12]. In this paper, we use the Takagi-sugeno fuzzy inference system. This system is composed of two inputs (the 2 components *Cb* and *Cr*) and one output (the decision skin or non skin color). Each input has three sub-sets: light, medium and dark. Our algorithm uses the concept of fuzzy logic IF-THEN rules; these rules are applied in each pixel in the image in order to decide whether the pixel represents a skin or non-skin region [24]. (Figure 3) represents the input image and the output one after *YCbCr* space conversion and fuzzy classification.

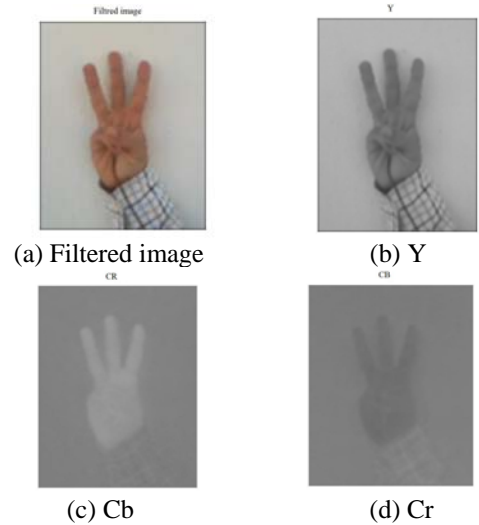

Figure 2.   Conversion of an *RGB* image into *YCbCr* space : (a) the original image, (b) the *Y* component, (c) the *Cb* component and (d) *Cr* component

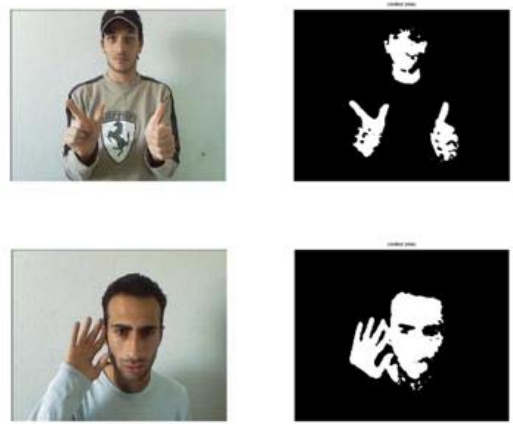

Figure 3.   Fuzzy classification of the image into skin regions and non skin regions

## B. Edge detection

Edge detection is the most common approach for detecting significant discontinuities in gray level. Edge detection algorithms locate and accentuate edges in order to identify objects in images.

Many different methods have been proposed for edge detection, such as Sobel filtering, Prewit filtering, Laplacian filtering, Moment based operator, Shen operator and Canny operator. Among the edge operators mentioned above, the most powerful edge-detection method is the Canny method. It differs from the other ones because it uses two different thresholds allowing it to detect strong and weak edges, and includes the weak edges in the output only if they are connected to strong edges [25]. In (figure 4), we show the robustness of Canny operator for detecting edges.

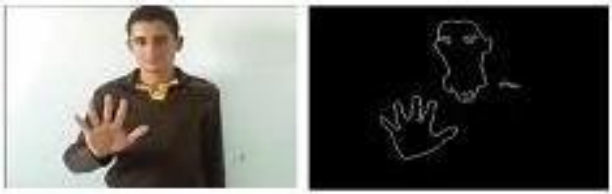

Figure 4.   Edge detection with Canny operators

After edge detection, the following step allowing to decide whether the region represents a hand or not.

## C. Hand localization

In this step we test two methods such as (Ellipse and Comparison based method) to localize hands.

*1)  Ellipse based method:* Because human face and hands have an elliptic shapes, an algorithm searching such shape is helpful for face and hand detection. Hough transformation is a well-known shape detection technique of image processing [26][27][28]. So, we use it in our detection algorithm.

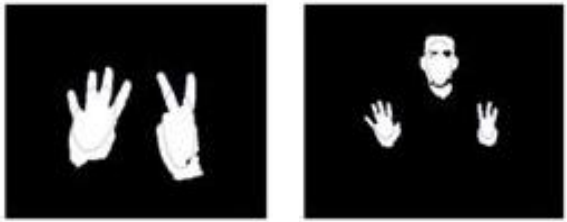

Figure 5.   A samples of ellipse visualization

As is known, face and hands have elliptical shapes, we must distinguish between them. To differentiate between the hands and the face, we must count the number of white pixels belonging to the perimeter of the ellipse. Then we calculate for each ellipse the ratio between the number of white pixels counted and the number of all pixels making up the perimeter. This ratio is small for the hand than for the face because the hand contains black regions between the fingers.

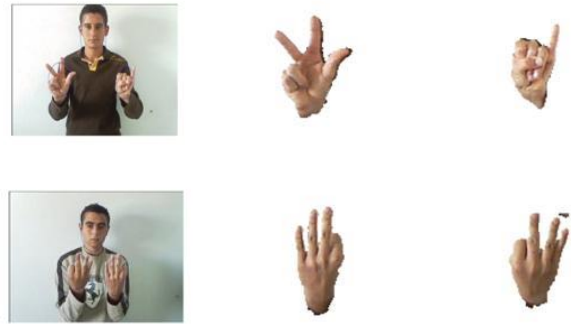

Figure 6.   Hand detection using elliptic method

*2)  Comparison based method:* This method consists to extract the three largest skin regions and to localize their barycenters. In order to distinguish the face from hands we propose the following rules depending on the number of skin regions detected in the image.

- Case 1: Image with three skin regions:

In this case we assume that there are a face and two hands in the image. Since the face have the largest area and the highest barycenter in the image, we can localize it and eliminate it. Each skin region is approximated by a rectangular surface characterized by its height and width in order to determine its area. The region having the largest area represents a potential face. To get the highest barycenter, we need to compare the three centers and extract who has the largest coordinate.

- Case 2: Image with two skin regions:

In this case we determine the ratio between the two surfaces of skin regions already detected. If the ratio is near to 1 (less than 1.5), we assume then that there is two hands in the image, else there are a hand and face. We must employ the same criteria (the largest area and the highest barycenter in the image) already mentioned to differentiate the face from the hand.

- Case 3: Image with a single skin region:

When having a unique skin region, we cannot apply any comparison and therefore we assume that it's a hand. This is the major weakness of this method.

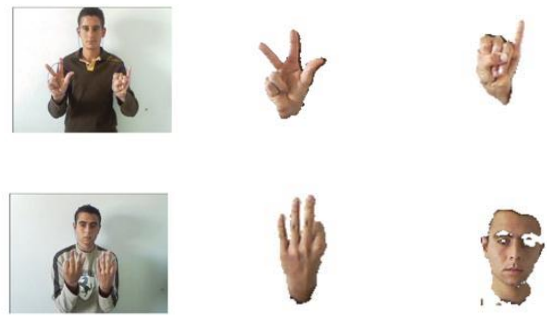

Figure 7.   Hand detection using comparison based method

*D. Hand adjustment into the vertical direction*

The hand adjustment is the most important phase in the hand localization steps. The main goal of this phase is to put the hand into the vertical direction (normalized hand direction in our approach). We give details of each phase in the following.

The human hands have an elliptic shape, an algorithm finding such shape is helpful for hand orientation.

The fit-estimation method we use is based on the least squares method. The input parameters to find the best ellipse are a set of *(x,y)* points (the *x,y* coordinates of all skin detected pixels), and the output is the minor and major axis, and the $\theta$ orientation for the best ellipse (the one that contains the majority of *(x,y)* input points). The former parameter is assumed as the hand orientation. (Figure 8) gives an illustration of ellipse determined to the hand shape. The $\theta$ orientation is given by the angle between the ellipse major axis and the *x* coordinate axis.

We use it to adjust the hand into the vertical direction.

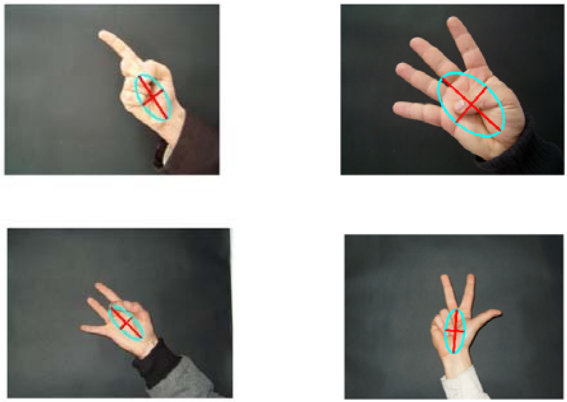

Figure 8. Ellipse representation for hand orientation

In order to obtain a best orientation we attempt to find an ellipse with major axis parallel to the hand axis. But when the thumb is far of other fingers we obtain an ellipse having a major axis not parallel to the hand one (figure 9). Thus we use morphology operations to resolve this problem to capture more precisely the elliptic shape.

The first morphology operation is dilation with a *diamond* structuring element to add a white pixels between all the fingers except the thumb (because the thumb is usually far from the other fingers), then we apply an erosion operation with a *disk* structuring element to eliminate a large part of the thumb. After these morphology operations, we put the rest of pixels coordinate in the ellipse finding algorithm.

Once the best ellipse is correctly found, we assume that the fingers belong to one of the two ellipse half's. We split the ellipse into two half's using the minor axis. The ellipse half that contains the fingers is the one that contains less skin pixels than the other; because of the gap that separates each pair of fingers.

Thus, in order to identify it, we compute the number of skin pixels in each ellipse half and retain the appropriate one.

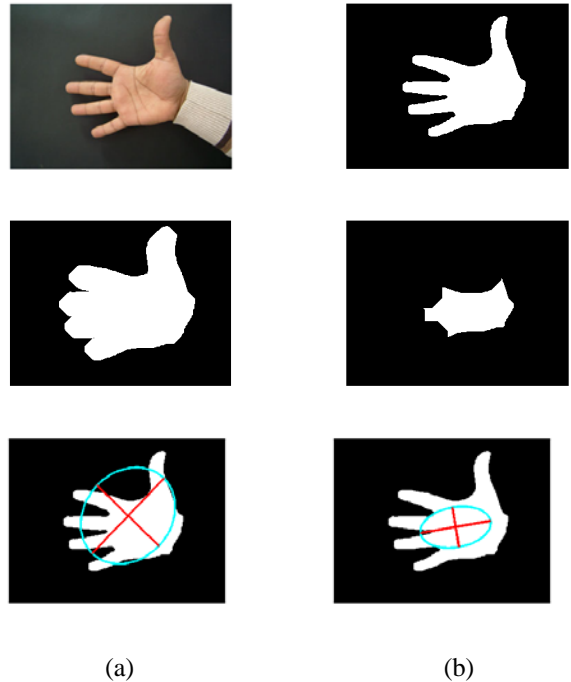

(a)                    (b)

Figure 9. Morphology operations correction : (a) finding ellipse without correction. (b) finding ellipse after correction

## III. FINGERS EXTRACTION

All the features used in the digit recognition phase are extracted from the active fingers attributes (localizations, distributions, heights and distances). Thus, we have to remove from the image all the hand components except the fingers. This step consists of two phases: palm localization and fingers extraction.

The palm localization phase is based on rules defined in the hand anthropometry study of comparative measurements of the human body [8]. These rules are useful to define the palm dimension, starting from hand dimensions:

- The palm length is equal to *0.496 ± 0.003* times the hand length.

- The palm width is equal to *0.44 ± 0.007* times the hand length.

To estimate the hand dimensions, we attempt to bound all skin pixels in the rectangular area having the smallest perimeter. The dimensions of this area will be considered as those of the hand[1]. We start by defining the convex hull of the skin pixels; the most appropriate rectangular area should contain at least one edge of the convex hull.

(Figure 10) provides the result of the hand bounding on a green rectangular area applied on some examples.

To localize the palm, we start by translating randomly a rectangle window having the palm dimensions.

---

1. These dimensions are different from the ellipse minor and major axis size, but give better estimation of hand dimensions

Then we localize it in the region having the maximum number of skin pixels. The region bounded by the red rectangle shown by (figure 10) is the palm region.

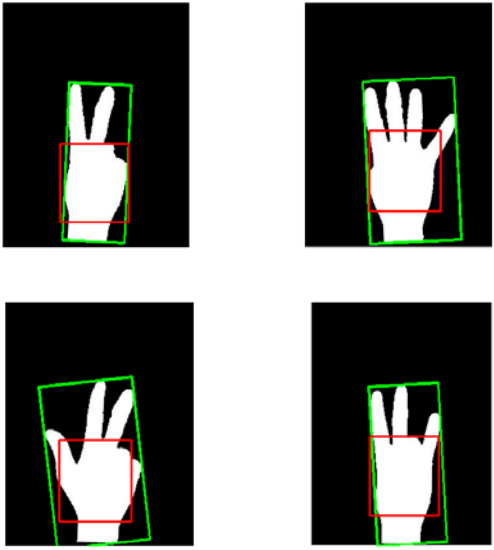

Figure 10. Palm localization

In the fingers extraction phase, we then attempt to remove from the image all the hand components except the fingers. First, we start by drawing a circle passing through the vertices of rectangle found in the previous phase, so we remove all the skin pixels in this circle. This process is made to remove the palm. The part of wist is eliminated by removing all the skin pixels below this rectangle.

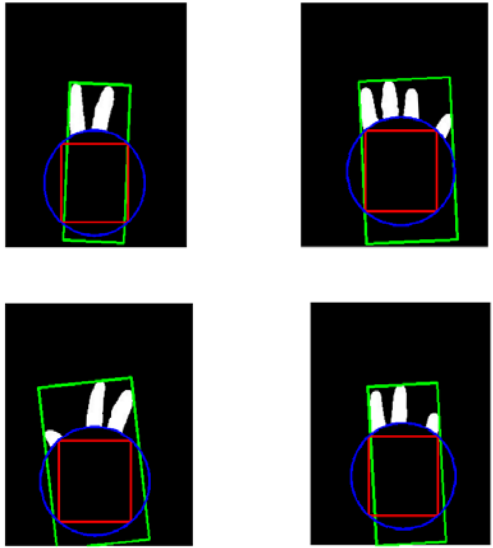

Figure 11. Removing hand components except fingers

## IV. FEATURES IDENTIFICATION FOR DIGIT RECOGNITION

The remaining skin pixels are those of the hand fingers. (Figure 11) shows that only the fingers pixels remain on the image.

Starting from these pixels, we project each one on the *x*-coordinate axis. For each *x* coordinate value *x*, we count the number of skin pixels *(x,.)*, the obtained numbers results on a histogram shown by (figure 12), with normalization by the hand length.

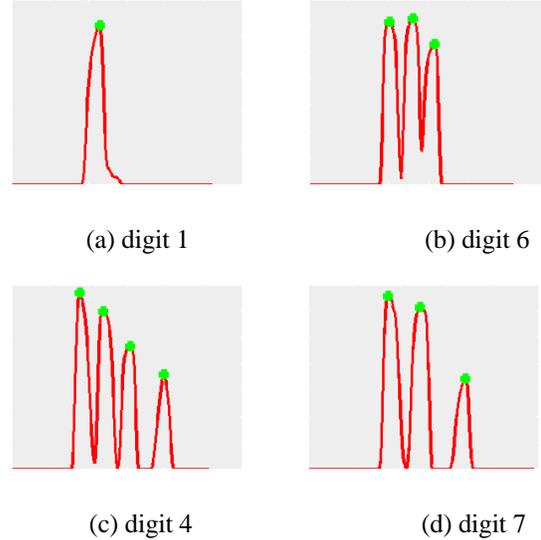

(a) digit 1          (b) digit 6

(c) digit 4          (d) digit 7

Figure 12. Histogram of skin pixels projection

The digit recognition is based on the set of peaks in the histogram. We have used the smoothing method to capture all the significant peaks in the histogram. The number of peaks provides a significant closes for the digit recognition. In fact, if the histogram contains one peak, the digit is *1* without any doubt. But if the number of fingers is *3*, many decisions could be made depending on other features. In such case, *3*, *6*, *7*, *8* and *9* are shown using *3* fingers, but fingers configurations are different from a digit to another. To differentiate between such cases, we have defined a set of additional features which take into account the distance between each pair of successive fingers and their lengths, which are done by the amplitudes of the retained peaks. Let us consider $(x_i,y_i)$, $i \in \{1...n\}$ the set of peaks where $n$ is the number of peaks (the number of active fingers), $x_i$ is the $x$ coordinate of the $i^{th}$ peak and $y_i$ is its amplitude. We compute the features $dist_{xi}=|x_i-x_{i+1}|$ and $dist_{yi}=|y_i-y_{i+1}|$, $i \in \{1...n-1\}$ and we attribute the zero value to $dist_{xi}$ and $dist_{yi}$ for all $n \leq i \leq 4$ when the number of peaks cannot exceed *5*. For example, if the histogram contains three peaks $(x_1,y_1)$, $(x_2,y_2)$ and $(x_3,y_3)$, we compute the distances $dist_{x1}=|x_1-x_2|$, $dist_{y1}=|y_1-y_2|$, $dist_{x2}=|x_2-x_3|$ and $dist_{y2}=|y_2-y_3|$, the remaining values of $dist_{x3}$, $dist_{y3}$, $dist_{x4}$ and $dist_{y4}$ are zeroed since they correspond to all the *i* values with $n=3 \leq i \leq 4$. These features are used to make our approach invariant of the horizontal and vertical positions of the hand in the image.

In order to enhance the invariance of our model to the hand scale, we compute an additional feature which represents the ratio between two successive non zero distances ($r_{xi}=dist_{xi}/dist_{xi+1}$ and $r_{yi}=dist_{yi}/dist_{yi+1}$). In fact, the distance ($dist_{xi}$, $dist_{yi}$) depends on the scale, but the ratio ($r_{xi}$, $r_{yi}$) corresponding to the same hand in two different scales is maintained the same. These features are then provided in a vector form, let:

$$V=(n, dist_{x1}, dist_{y1}...dist_{x4}, dist_{y4}, r_{x1}, r_{y1}... r_{x4}, r_{y4})$$

where $n$ is the number of detected peaks. A set of vectors in such format is provided as input of the decision tree based algorithm which looks for the best decision rules to apply in order to decide which digit is shown. The algorithm is based on a set of learning vectors to identify these rules.

## V. EXPERIMENTS AND RESULTS

Our system has been evaluated using two series of tests. In the first one, we have used a dataset containing 920 images; each one contains only one person to evaluate our detection and localization system. In the second one, a dataset containing 1980 hand images has been used to evaluate our digit recognition system. These images have different sizes, lighting conditions and complex backgrounds. A study of performance of each part of our system (skin color detection, hand detection and digit recognition) is detailed in this section.

### A. Skin color detection

We calculate the detection ratio in two levels:

- After classic classification using arrangements in *Cb* and *Cr* plane.

- After fuzzy classification.

Results are shown in (table 1).

TABLE I. THE CLASSIFICATION STEP RESULTS

| Detection of skin region | Skin color detection ratio |
|---|---|
| Classic classification | 91.67% |
| Fuzzy Classification | 96.4% |

The standard classification has detected 1327 hands from 1448 total hands, while the fuzzy classification has detected 1396 hands. Consequently we opt for fuzzy classification because it is more efficient in skin regions detection rate.

### B. Hand detection

(Table 2) shows the performances of the two methods used for hand detection step such as, the Comparison based method and the ellipse method.

TABLE II. THE GESTURE HAND RECOGNITION RESULTS

| Detection of hands | Hand detection ratio |
|---|---|
| Comparison based method | 90,4% |
| Ellipse method | 93,7% |

The Comparison based method has been detected 1262 hands from 1396 detected in the previous classification step. But the ellipse method has been detected 1308 hands. According to these observations, the ellipse method is the better for our application. So we will use its results in the next step to recognize hand gestures.

### C. Digit recognition

This database has been divided randomly into a set of training images (*70%*) and a set of test images (the remaining *30%*). In the literature, there are several techniques of supervised learning. We have evaluated three techniques of graphs of decision tree [14] to know: ID3 [12], C4.5 [13] and IMPROVED C4.5 [15].

We present two series of experiments: in the first, we experimented with the three techniques mentioned above on the training data set and validate the quality of the learned models using random error rate techniques. Three measures have been used: *precision rate*, *recall rate* for each digit and *classification accuracy rate*.

The performance of obtained classifiers can be assessed by a confusion matrix opposing assigned class (column) of the samples by the classifier with their true original class (row). (Table 3) illustrate, the confusion matrix used for the validation stage. Three global indicators on the quality of a classifier from such a confusion matrix can be build:

TABLE III. CONFUSION MATRIX

$$\begin{pmatrix} Class & 1 & 2 & \cdots & 9 \\ 1 & N_{1,1} & N_{1,2} & \cdots & N_{1,9} \\ 2 & N_{2,1} & N_{2,2} & \cdots & N_{2,9} \\ \vdots & \vdots & \vdots & \ddots & \vdots \\ 9 & N_{9,1} & N_{9,2} & \cdots & N_{9,9} \end{pmatrix}$$

- Global error rate: is the complement of classification accuracy rate or success classification rate.

$$\varepsilon_{global} = \sum_{1 \le i, j \le 9; i \ne j} \frac{N_{i,j}}{card(M)}$$

Where *card(M)* is the number of samples in a test bed.

- A priori error rate: this indicator measures the probability that a simple of class (*Digit_i*) where $I \in \{1...9\}$ is classified by the system to other class.

$$\varepsilon_{apriori}(Digit_i) = \frac{\sum_{j=1, j \ne i}^{9} N_{i,j}}{\sum_{j=1}^{9} N_{i,j}}$$

This indicator is thus clearly the complement of the classical recall rate which is defined for ($Digit_i$) where $i \in \{1...9\}$ class by:

$$\mathrm{Re}\,call(Digit_i) = \frac{N_{i,i}}{\sum\limits_{j=1}^{9} N_{i,j}}$$

- A posteriori error rate: this indicator measures the probability that a simple assigned to class ($Digit_i$) where $i \in \{1...9\}$ by the system effectively belongs to class ($Digit_i$) where $i \in \{1...9\}$.

$$\varepsilon_{aposteriori}(Digit_i) = \frac{\sum\limits_{j=1, j \neq i}^{9} N_{j,i}}{\sum\limits_{j=1}^{9} N_{j,i}}$$

This indicator is thus clearly the complement of the classical precision rate which is defined for ($Digit_i$) where $i \in \{1...9\}$ class by:

$$\mathrm{Pr}\,ecision(Digit_i) = \frac{N_{i,i}}{\sum\limits_{j=1}^{9} N_{j,i}}$$

A good decision tree obtained by a data mining algorithm from the learning dataset should not only produce classification performance on data already seen but also on unseen data as well. In our experiments, in order to ensure the performance stability of our learned models from the learning data, we thus also tested the learned models on our test data.

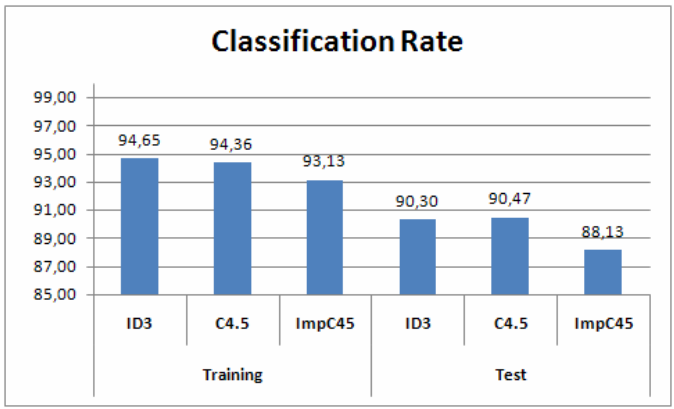

Figure 13. Classification rate

(Figure 13) shows that our approach provides very satisfying results over *90%* in our test dataset on the classification accuracy rate, and that ID3 and C4.5 provide the best classification rates. As for us, we retain ID3.

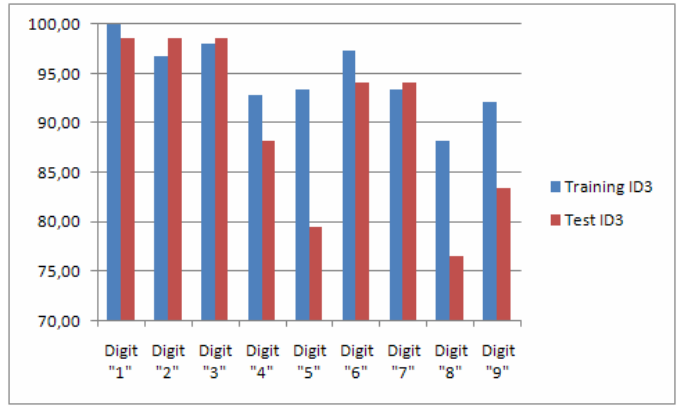

Figure 14. Recall rate (ID3 classifier)

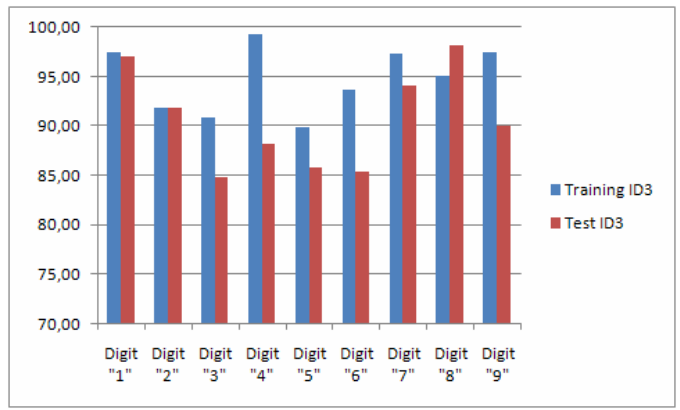

Figure 15. Precision rate (ID3 classifier)

We emphasize the remaining of our experiments on the recall and precision rates obtained for each digit. Over our experiments, we show that our system provides a good classification rate on some digits, in particular for digits *1* and *2*. In fact, according to these digits, the number of peaks is still sufficient for the decision rule, indeed, the classifier ID3 (as well as the others) puts the emphasis on these criteria.

The confusion problem, which causes less performance, is shown in the other digits; we observed that the recall rate of digit *4* is relatively low because of many false alarm cases of digit *5*.

This confusion problem is basically caused by the thumb position which is sometimes plotted with the index finger, so that, only *4* peaks are detected.

The second confusion problem is observed for digits *3,6,7,8* and *9*, where exactly *3* peaks are detected (not that the thumb is not considered in digits *6,7,8* and *9*, and in digit *3*, the thumb is correctly plotted in general).

Several precision and recall rates are shown because of the same problem of false alarm. The main cause of this problem is the distance between the fingers used to sign each digit which are different from a person to another.

In our future work, we think of include some other aposteriori criteria's to differentiate between digits *4* and *5* when *4* peaks are detected, and an additional study of the fingers distances in the case of digits *3,6,7,8* and *9*.

## VI. CONCLUSION

We have presented in this paper an approach of digit recognition based on fingers detection and analysis. The analysis requires many adjustment and removal steps to avoid inefficient skin pixels influence. On the other hand, the human hand properties are straight forward for the sign language. In fact, thanks to these properties many measures have been correctly estimated and used in several steps of our approach.

Throw on experiments, we show that the approach is effective and the performance is over *90%* in our test data set as classification accuracy rate. Our future orientation concerns the use of other hand properties, such as finger length. We also think about identifying alphabetic signs.